\documentclass[letterpaper, 10 pt, conference]{ieeeconf}
\IEEEoverridecommandlockouts
\usepackage{subcaption}
\usepackage{amsmath,amssymb}
\usepackage{graphicx}
\usepackage{booktabs}
\let\labelindent\relax
\usepackage{enumitem}
\usepackage{hyperref}
\usepackage{xcolor}
\usepackage{titlesec}
\usepackage{caption}
\usepackage{setspace}
\usepackage{tabularx}
\usepackage{parskip}
\usepackage{float}
\usepackage{cite}
\usepackage{algorithm}
\usepackage{algpseudocode}
\usepackage{stfloats}

\renewcommand{\baselinestretch}{1.0}
\usepackage{tikz}
\usetikzlibrary{arrows.meta,decorations.markings,calc}
\definecolor{titleblue}{HTML}{1F4E79}
\definecolor{sectionblue}{HTML}{2E75B6}

\hypersetup{
    colorlinks=true,
    linkcolor=sectionblue,
    citecolor=sectionblue,
    urlcolor=sectionblue
}

\begin{document}
%
\title{
HINT-Blimp: Human INTent Inference from Multimodal Cues for Robotic Blimps
}
%
%
%

\author{Subhadeep Koley$^{1}$, Benjamin Greenberg$^{2}$,
Yifei Simon Shao$^{3}$,
Juan Aceros$^{1}$, Nadia Figueroa$^{3}$, David Salda\~{n}a$^{1}$
\thanks{$^{1}$Subhadeep Koley, Juan Aceros and David Salda\~{n}a
are with Lehigh University, Bethlehem, PA 18015, USA.
{\ttfamily \{svk324,\allowbreak saldana\}@lehigh.edu}}
\thanks{$^{2}$Benjamin Greenberg is with Swarthmore College,
Swarthmore, PA 19081, USA.
\nolinkurl{bgreenb2@swarthmore.edu}}
\thanks{$^{3}$Yifei Simon Shao and Nadia Figueroa are with
the University of Pennsylvania, Philadelphia, PA 19104, USA.
{\ttfamily \{yishao,\allowbreak nadiafig\}@seas.upenn.edu}}
}

\markboth{Journal of \LaTeX\ Class Files,~Vol.~14, No.~8, August~2015}%
{Shell \MakeLowercase{\textit{et al.}}: Bare Demo of IEEEtran.cls for IEEE Journals}

\maketitle
\begin{abstract}
In human-robot interaction, traditional interfaces such as joysticks and handheld tablets introduce latency into navigation tasks and require the operator's explicit attention on the device, instead of the robot. We propose a new human-robot interaction framework in which a human communicates intent directly through sparse multimodal signals such as physical pushes and spoken commands. Human intent is represented as a parameterized linear dynamical system (LDS) that encodes
the desired goal and motion behavior. The robot estimates this
intent (parameters) online using a particle filter, where each
particle represents a candidate LDS hypothesis and is reweighted
online as new information becomes available. We validate this
framework on a robotic blimp, whose inherent compliance and
collision tolerance make it well-suited for repeated physical
interaction. Experiments with multiple participants across 300 trials show that combining pushes and voice commands identifies the intended goal in 86\% of trials within at most five interactions, with most trials resolved in two. The inferred dynamical systems can also produce curved trajectories that
avoid obstacles known only to the human.
Video : \url{https://youtu.be/b_QbHGdgBf4}

\end{abstract}

\begin{figure}[t]
    \centering
    \includegraphics[width=1.0\linewidth]{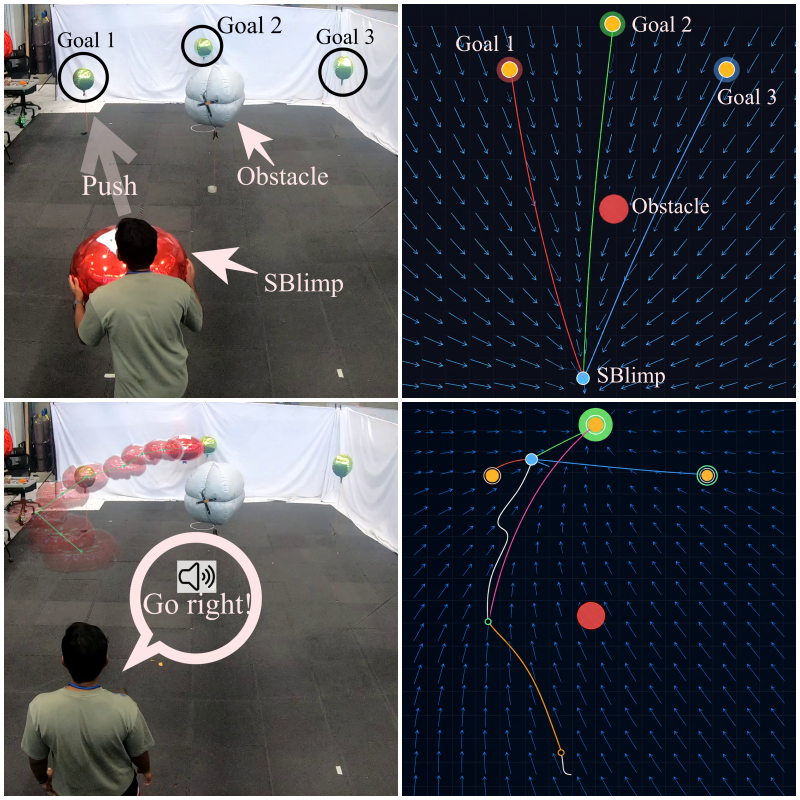}
    \caption{HINT-Blimp infers human intent from multimodal interactions. Human–robot interaction occurs exclusively through physical pushes and voice commands.
    In this scenario, the user intends to send the blimp to Goal 2 but cannot push it directly toward the goal because a straight-line trajectory would collide with the obstacle. Instead, the user first pushes the blimp toward Goal 1 and then redirects its motion using the verbal command ``Go right!''. Our probabilistic method (right column) maintains beliefs over candidate trajectories to each goal and progressively converges to the correct intent as additional information is provided by the user.}
    \label{fig:comp_tile.png}
\end{figure}

\IEEEpeerreviewmaketitle

\section{Introduction}

{Understanding} human intent is necessary for robots to operate in shared human-robot environments. During robot navigation tasks, humans often need to correct, redirect, or refine a robot's behavior without stopping it. In such cases, conventional communication channels, such as graphical user interfaces with goal-selection menus, can introduce latency and require explicit attention from the human. Physical and verbal cues provide a more immediate alternative: a brief push, spoken command, or pointing gesture can convey information about the desired goal or behavior without requiring the human to operate an external device or interface. Interpreting such cues is challenging because they are short, sparse, and often ambiguous. The same push direction may be consistent with multiple goals or multiple motion behaviors, especially in cluttered environments or when the desired trajectory is not a straight line to the goal. Similarly, a spoken command such as ``go left'' may not specify a complete trajectory, but can still provide useful evidence that disambiguates among candidate goals. This motivates an inference framework that does not treat human input as a complete command, but instead uses each cue as partial evidence about the human's intended goal and trajectory.


Intent inference in physical human-robot interaction has been widely studied for robotic manipulators in co-manipulation and shared-control tasks~\cite{losey2018review_intent_shared_control,human_intent_in_comanipulation,var_impedence}. Dynamical systems have been particularly useful in these settings because they provide smooth and stable motion representations for physical interaction and task adaptation~\cite{billard2022learning,khoramshahi2019ds_task_adaptation,khoramshahi2020ds_detection}. However, expressing high-level intent through sparse physical cues remains relatively underexplored for mobile robots. Prior work on physical interaction with wheeled and aerial robotic platforms has primarily focused on low-level admittance, impedance, or force-estimation-based control~\cite{wheeled_ballbot,admitance_quad,franchi_phri,guiseppe_collaboration,marco_nmpc}. Some works have also explored tactile and touch-based human-drone interaction, such as virtual touch buttons on quadrotors~\cite{metrodrone}, safe-to-touch drone interfaces~\cite{abtahi2017touch_drone}, and physical interaction through direct contact or tether forces~\cite{banks2021physical_human_uav,tognon2021tethered,allenspach2022human_state_aware,hallworth2023state_aware}.

While these approaches demonstrate the feasibility of close physical interaction with aerial robots, they primarily treat touch or force as a compliant control input rather than as evidence for high-level intent. We address this gap using aerial blimps, whose low-speed dynamics, inherent compliance, and collision tolerance make them well-suited for repeated physical interaction in shared environments~\cite{xu2023sblimp}.

The main contribution of this work is a new multimodal intent-inference framework for mobile robots. Intent is represented jointly by a candidate goal and the parameters of a dynamical-system motion policy, allowing cues to provide information about \emph{where} the human wants the robot to go and \emph{how} it should get there. Physical pushes and spoken commands are incorporated as complementary sources of evidence within a unified Bayesian estimation problem. This belief over the user's intent is maintained online using a particle filter, enabling the robot to preserve uncertainty over candidate goals and motion behaviors and refine its estimate as new cues become available. 
We validate the framework on a physical blimp in a study with multiple participants, showing that fusing multimodal cues achieves 86\% goal identification and that the inferred dynamical systems can produce curved trajectories that avoid obstacles known only to the human.

\section{Preliminaries}
\label{lds}

\begin{figure}[h]
    \centering
    \includegraphics[width=1.0\linewidth]{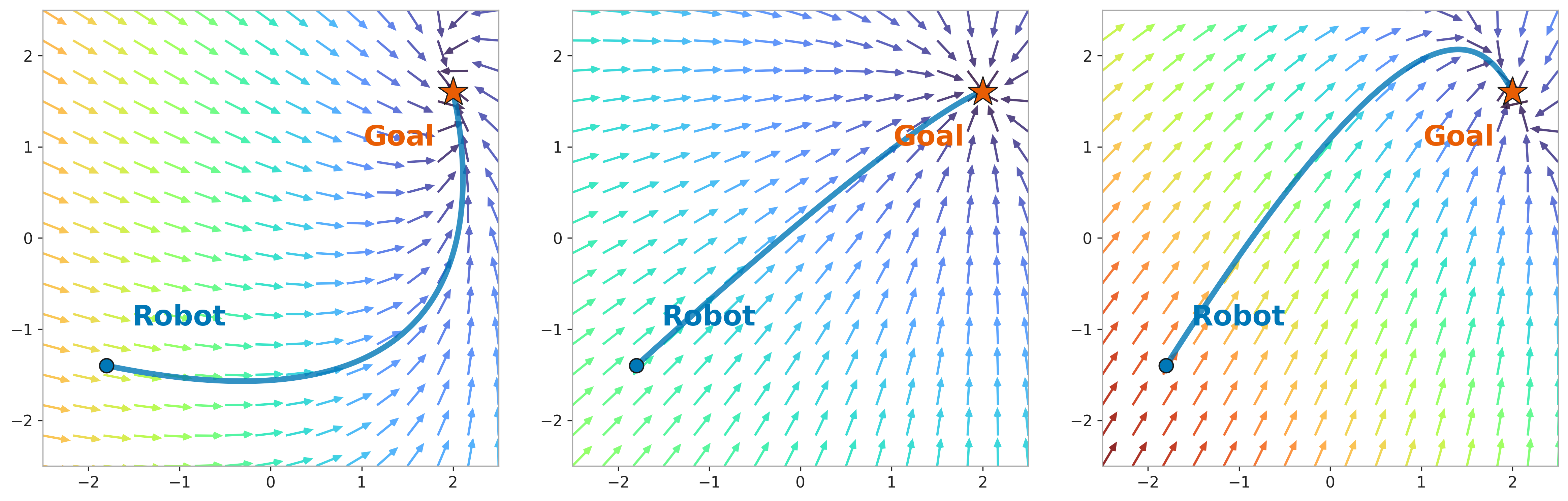}
    \caption{Examples for vector fields and robot trajectories for LDS. From left to right, the LDS parameters are $(\vartheta_1,\vartheta_2,r)=(0.58,103^\circ,133^\circ)$, $(0.90,117^\circ,70^\circ)$, and $(0.47,117^\circ,70^\circ)$.}
    \label{fig:lds}
\end{figure}

We use linear dynamical systems (LDS) to represent desired robot motion. For a robot operating in $\mathbb{R}^2$, we consider a dynamical system of the form
\[
\dot{\mathbf{x}} = f(\mathbf{x};\, \boldsymbol{\theta}),
\]
where $\mathbf{x} \in \mathbb{R}^2$ is the robot position and $f$ defines a velocity field parameterized by $\boldsymbol{\theta}$. For goal-directed motion, we use the LDS
\begin{equation}
\dot{\mathbf{x}} = \mathbf{A}(\mathbf{x} - \mathbf{x}^*),
\label{eq:lds}
\end{equation}
where $\mathbf{x}^* \in \mathbb{R}^2$ is the position of the goal and $\mathbf{A} \in \mathbb{R}^{2 \times 2}$ has real, strictly negative eigenvalues, guaranteeing global asymptotic stability at $\mathbf{x}^*$. In this linear case, the intent parameters are $\boldsymbol{\theta} = \{\mathbf{A},\, \mathbf{x}^*\}$, comprising the four entries of $\mathbf{A}$ and the two components of the goal position. The eigenstructure of $\mathbf{A}$ determines the qualitative behavior of the motion: isotropic contraction produces straight-line convergence, while different contraction rates along the eigendirections can produce curved trajectories (Fig.~\ref{fig:lds}).  More general eigenstructures of $\mathbf{A}$ can produce other motion behaviors, such as spiraling or orbiting around an equilibrium. In this paper, we restrict $\mathbf{A}$ to the stable-node case.

\section{Problem Statement}
\label{sec:problem}

We consider a scenario where a human wishes to send a robotic blimp to one of $K$
candidate goal locations $\{\mathbf{x}^*_i\}_{i=1}^{K}$. The human
communicates the intent (desired goal and behavior) through sparse,
multimodal interactions, and the robot's objective is to identify the
intended goal and execute the corresponding motion by controlling the vehicle's velocity. Throughout, we use
$\mathbf{x} \in \mathbb{R}^2$ and $\psi_{\text{robot}} \in S^1$ to denote
the blimp's position and heading in the world frame at any given time.

\paragraph{Intent as a linear dynamical system.}
The human intent is modeled as an LDS (from \eqref{eq:lds}), and our method will focus on inferring its parameters $\boldsymbol{\theta} = (\mathbf{x}^*,
\mathbf{A})$, where $\mathbf{x}^*$ encodes \emph{where} the intended goal is and
$\mathbf{A}$ encodes \emph{how} the robot should reach it. 

\paragraph{Input modalities.}
The robot receives evidence from any of the following inputs.

\begin{itemize}
    \item \textbf{Scene context} $\mathcal{S} = \{(\mathbf{x}^*_i,\, \pi_i)\}_{i=1}^{K}$:
    The scene provides a set of $K$ candidate goal locations with associated prior plausibilities
    $\pi_i \geq 0$, $\sum_i \pi_i = 1$. Unlike the other modalities, it does
    not arise from direct human interaction but instead encodes background
    knowledge about which destinations are relevant to reach. 

\item \textbf{Physical pushes.} The human pushes the blimp in a
particular direction, and the resulting short burst of motion carries
information about the intended goal location. We capture this as a
set of $N_p$ push observations
$\mathcal{Z}_{\text{push}} = \{\mathbf{z}^{\text{push}}_j\}_{j=1}^{N_p}$,
where each observation is a trajectory segment
\begin{equation}
    \mathbf{z}^{\text{push}}_j = \mathcal{D}_\tau^{(j)}
    = \bigl\{\mathbf{x}(t),\,\dot{\mathbf{x}}(t)\bigr\}_{t=t_j}^{t_j+\tau},
    \label{eq:push_obs}
\end{equation}
recorded over a time window of length $\tau$.

    \item \textbf{Directional hints.} The human can also
    redirect the blimp without touching it, by saying ``go left'' or
    pointing in the intended direction. We treat speech and gesture as
    two sources of the same kind of evidence, a direction vector relative to the
    blimp's current heading, and collect them as a set of $N_d$
    directional observations, where $\mathcal{Z}_{\text{dir}} =
    \{\mathbf{z}^{\text{dir}}_j\}_{j=1}^{N_d}$ and 
    \begin{equation*}
        \mathbf{z}^{\text{dir}}_j = \bigl(\text{source}_j,\, \mathbf{u}_j\bigr),
        \qquad \text{source}_j \in \{\text{voice},\, \text{gesture}\},
        \label{eq:dir_obs}
    \end{equation*}
    each carrying a source tag and a directional payload
       \begin{equation}
        \mathbf{u}_j = \bigl(\mathbf{d}^{\text{cmd}}_j,\,
        \mathbf{x}_{\text{robot},j},\,
        \psi_{\text{robot},j}\bigr),
        \label{eq:dir_payload}
    \end{equation}
    that records a commanded direction vector
    $\mathbf{d}^{\text{cmd}}_j \in S^1$ together with the blimp's
    position and heading at command time. The vector is expressed in
    the robot's heading frame, so that $(1,0)$ means straight ahead
    and $(0,1)$ means right.

\end{itemize}

The human interactions arrive as an ordered sequence of observations
$\mathbf{z}_{1:k} = (\mathbf{z}_1, \mathbf{z}_2, \dots, \mathbf{z}_k)$, where
each $\mathbf{z}_k \in \mathcal{Z}_{\text{push}} \cup \mathcal{Z}_{\text{dir}}$
may be a physical push or a voice command or a gesture, arriving in any order and at any
time. The robot's objective is to infer, from this sequence, the intended
goal~$\mathbf{x}^*$ and behavior $\mathbf{A}$ that jointly define the
dynamical system $\dot{\mathbf{x}} = \mathbf{A}(\mathbf{x} - \mathbf{x}^*)$.
The robot should then follow the estimated LDS.

\section{Inferring Human Intent}
\label{methodology}

Our goal is to infer the human's intent, parameterized by the parameters 
$\boldsymbol{\theta}$, from a sequence of observations $\mathbf{z}_{1:k}$.
We design a particle filter that estimates the human intent incrementally as each new observation arrives.

\subsection{Eigenvalue Parameterization}
\label{eig}

Rather than estimating the four entries of $\mathbf{A} \in \mathbb{R}^{2 \times 2}$ directly, we parameterize it through its eigen-decomposition. Since we require a stable node (real, negative eigenvalues), we fix one eigenvalue at $\lambda_1 = -1$ and parameterize the relative contraction rate $r = |\lambda_1|/|\lambda_2| \in (0,1]$, so that $\lambda_2 = -1/r$.  

In the general case, the matrix is as:
\begin{equation}
\label{eq:reconstruct}
\mathbf{A} = \mathbf{V}\,\mathrm{diag}(-1,\, -1/r)\,\mathbf{V}^{-1},
\end{equation}
where $\mathbf{V} = [\mathbf{v}(\vartheta_1) \quad \mathbf{v}(\vartheta_2)]$ is formed from unit eigenvectors at angles $\vartheta_1, \vartheta_2 \in [0, \pi)$. Specifically, each eigenvector is written as
\begin{equation*}
\mathbf{v}(\vartheta)
=
\begin{bmatrix}
\cos \vartheta \\
\sin \vartheta
\end{bmatrix},
\text{so that }
\mathbf{V}
=
\begin{bmatrix}
\cos \vartheta_1 & \cos \vartheta_2 \\
\sin \vartheta_1 & \sin \vartheta_2
\end{bmatrix}.
\end{equation*}

A minimum angular separation between $\vartheta_1$ and $\vartheta_2$ keeps $\mathbf{V}$ invertible and avoids nearly parallel eigendirections. Thus, each particle is re-parameterized as $\boldsymbol{\theta} = (\vartheta_1,\, \vartheta_2,\, r,\, \mathbf{x}^*)$, instead of the original $(\mathbf{x}^*, \mathbf{A})$ of the problem statement. Compared to estimating the entries of $\mathbf{A}$ directly, this gives a lower-dimensional and more structured search space, while explicitly enforcing stability, controlling relative contraction rates, and avoiding degenerate eigenvector configurations.

\subsection{Particle Filter for Intent Estimation}

We represent the robot's belief over the human's intent as a
probability distribution $p(\boldsymbol{\theta}_k \mid \mathbf{z}_{1:k})$. We use a set of $N$ weighted particles
$\{\boldsymbol{\theta}_k^{(n)},\, w_k^{(n)}\}_{n=1}^N$, where
$\boldsymbol{\theta}_k^{(n)}$ denotes the $n$-th candidate intent
hypothesis and $w_k^{(n)}$ its normalized importance weight at step
$k$. The classical particle filter~\cite{thrun2005probabilistic} maintains this belief by
iterating over three steps: sampling from a process model; computing
importance weights from an observation likelihood; and resampling based on particle weights. In our setting, the state is a parameterized LDS; intent is assumed to be static between
interactions, so the process model has no control input and reduces to
a near-static random walk; the observation likelihood is
modality-dependent; and the particle distribution is initialized using a scene prior over candidate goals. The full procedure is given in Algorithm~\ref{alg:intent_pf}.
 
\smallskip

\subsubsection{Process Model (Line~\ref{line:predict})}
\label{process_model}
Since intent does not change between interactions, we assume a
near-static evolution:
\begin{equation}
\label{eq:process}
\boldsymbol{\theta}_k^{(n)} = \boldsymbol{\theta}_{k-1}^{(n)}
                            + \boldsymbol{\eta}_k^{(n)},
\end{equation}
where $\boldsymbol{\eta}_k^{(n)}$ is zero-mean Gaussian noise applied
independently to the shape components $(\vartheta_1, \vartheta_2, r)$ and
scaled by $(1 - \varsigma_{k-1})$, with $\varsigma_{k-1} \in [0,1]$ a push-dependent
confidence computed after the previous filter step(discussed at the end of Section~\ref{physical_pushes}). The goal
component is held fixed at the candidate value assigned at initialization, so
$\boldsymbol{\eta}_k^{(n)}$ has no $\mathbf{x}^*$ entry: goals form a discrete
hypothesis set, and the belief over them evolves only through reweighting and
resampling.

\smallskip
\subsubsection{Observation Likelihood (Line~\ref{line:lik})}

Observations from any modality arrive sequentially and are denoted
$\mathbf{z}_{1:k}$, where each
\begin{equation}
    \mathbf{z}_k = (\text{type}_k,\, \text{payload}_k), \qquad
    \text{type}_k \in \{\text{push},\, \text{dir}\}
\end{equation}
carries a type tag and a payload: either a trajectory segment
$\mathcal{D}_\tau^{(j)}$ from~\eqref{eq:push_obs} when
$\text{type}_k = \text{push}$, or a directional tuple
$\mathbf{u}_j$ from~\eqref{eq:dir_payload} when
$\text{type}_k = \text{dir}$. Directional observations
further carry a source tag
$\text{source}_k \in \{\text{voice},\, \text{gesture}\}$
identifying whether the hint came from a spoken command or
a pointing gesture, though both are processed identically.

\begin{algorithm}[t]
\caption{\textsc{Intent\_Particle\_Filter}$(\mathcal{X}_{k-1},\, \mathbf{z}_k,\, \mathcal{S})$}
\label{alg:intent_pf}
\begin{algorithmic}[1]
\Require Particle set
         $\mathcal{X}_{k-1} = \{\boldsymbol{\theta}_{k-1}^{(n)}\}_{n=1}^N$,
         observation $\mathbf{z}_k = (\text{type}_k,\, \text{payload}_k)$,
         scene context $\mathcal{S}$
\Ensure Updated set $\mathcal{X}_k$, intent estimate $\boldsymbol{\theta}_k^*$
\Statex
\For{$n = 1$ to $N$}
    \State $\boldsymbol{\theta}^{(n)} \gets \boldsymbol{\theta}_{k-1}^{(n)}
            + \boldsymbol{\eta}^{(n)}$
        \Comment{process model~\eqref{eq:process}} \label{line:predict}
    \State $w^{(n)} \gets p\bigl(\mathbf{z}_k \mid \boldsymbol{\theta}^{(n)}\bigr)$
        \Comment{push~\eqref{eq:push_lik} or directional~\eqref{eq:voice_lik}}
        \label{line:lik}
\EndFor
\State $w^{(n)} \gets w^{(n)} \big/ \textstyle\sum_m w^{(m)}
       \quad \forall\, n$
    \Comment{normalize}
\State $i_k^* \gets$ goal candidate with the largest total weight
    \label{line:map}
\State $\boldsymbol{\theta}_k^* \gets$ highest-weight particle allocated to $i_k^*$
\State $\mathcal{X}_k \gets$ resample  particles 
 with probability
       $\propto w^{(n)}$ \label{line:resample}
\State \Return $\mathcal{X}_k,\, \boldsymbol{\theta}_k^*$
\end{algorithmic}
\vspace{0.3em}\hrule\vspace{0.4em}
{\footnotesize Each particle
$\boldsymbol{\theta}^{(n)} = (\vartheta_1^{(n)},\, \vartheta_2^{(n)},\,
r^{(n)},\, \mathbf{x}^{*(n)})$ is the hypothesis of a LDS with
importance weight $w^{(n)}$, and $\boldsymbol{\eta}^{(n)}$ is zero-mean
Gaussian process noise on the shape components only. Particles are
initialized by sampling goals from the scene prior $\mathcal{S}$ and, since
goals are static, each particle stays assigned to one candidate goal
throughout. }
\end{algorithm}

\noindent\textbf{Scene context for initialization:}
\label{scene_context}
The filter is initialized from the scene context
$\mathcal{S} = \{(\mathbf{x}^*_i,\, \pi_i)\}_{i=1}^{K}$, where $K$ is the number
of candidate goals and $\pi_i$ is the prior plausibility of goal
$\mathbf{x}^*_i$. Particles are sampled in proportion to these weights, so goals
that are more likely to be the intended target attract more particles from the
start. The prior encodes knowledge available before any human cue is observed,
and its source is left open: it may come from a semantic scene understanding
module (e.g a Vision Language Model), from task context such as the time of day, or from an operator
specifying candidate destinations directly. We treat $\pi_i$ as given and focus
on how sparse human cues reshape this prior into a posterior over intent.

\noindent\textbf{Physical pushes:}
\label{physical_pushes}
Physical pushes provide direct kinematic evidence about the user's
intended goal. The $j$-th push is recorded as the trajectory segment
$\mathcal{D}_\tau^{(j)}$ of~\eqref{eq:push_obs}, but the segment is not
available all at once. Its samples arrive continuously over the window
$\tau$ while the robot is still moving due to the human's imparted impulse.
The filter therefore consumes the push incrementally. Writing $t_k$ for the
time of filter step $k$, let
\begin{equation}
  \mathcal{W}_k^{(j)} = \bigl\{(\mathbf{x}_t,\, \dot{\mathbf{x}}_t)
  \in \mathcal{D}_\tau^{(j)} \;:\; t_{k-1} < t \le t_k \bigr\}
  \label{eq:increment}
\end{equation}
denote the samples recorded since the previous filter step. The increments
are disjoint and $\bigcup_k \mathcal{W}_k^{(j)} = \mathcal{D}_\tau^{(j)}$,
so each sample belongs to exactly one $\mathcal{W}_k$. At step $k$ the filter
scores only $\mathcal{W}_k^{(j)}$, through the summed cosine similarity
between observed and predicted velocities:
\begin{equation}
\label{equation:alignment}
  c_k^{(n)} = \sum_{t \in \mathcal{W}_k^{(j)}}
  \frac{\dot{\mathbf{x}}_t \cdot \hat{\dot{\mathbf{x}}}_{t}^{(n)}}
       {\|\dot{\mathbf{x}}_t\|\,\|\hat{\dot{\mathbf{x}}}_{t}^{(n)}\|},
\end{equation}
where $\hat{\dot{\mathbf{x}}}_{t}^{(n)} =
\mathbf{A}^{(n)}(\mathbf{x}_t - \mathbf{x}^{*(n)})$ is the velocity
predicted by particle $n$'s dynamical system at the observed position
$\mathbf{x}_t$. This purely directional scoring is consistent with the
speed-invariant parameterization of the LDS. Motion capture arrives at a
known rate $f_m$, so a window of length $\tau$ yields $N_\tau = \tau f_m$
samples, a quantity known when the push is detected. Normalizing by it
gives the likelihood
\begin{equation}
\label{eq:push_lik}
  p\!\left(\mathcal{W}_k^{(j)} \mid \boldsymbol{\theta}^{(n)}\right)
  \propto \exp\!\Bigl(\tfrac{\kappa_p}{N_\tau}\, c_k^{(n)}\Bigr),
  \qquad \kappa_p > 0 .
\end{equation}
Each increment thus spends the fraction $|\mathcal{W}_k^{(j)}| / N_\tau$ of
a fixed budget $\kappa_p$, and since the increments together cover the entire push segment without overlap, these fractions sum to one and the total log-likelihood contributed over the push is $\kappa_p\,\bar{c}_j^{(n)}$, where $\bar{c}_j^{(n)}$ is the mean cosine similarity over the complete segment. 

The confidence $\varsigma_k \in [0,1]$ is computed after the update at step $k$
and scales the process noise at step $k+1$ by $1-\varsigma_k$.
It is the mean cosine alignment over all push samples received so far,
evaluated at the current estimate $\boldsymbol{\theta}_k^*$ and floored at
zero. Specifically, it is obtained by applying
\eqref{equation:alignment} to
$\bigcup_{i \le k}\mathcal{W}_i^{(j)}$
and dividing by the number of samples. A well-explained push therefore
suppresses subsequent exploration, while a poorly explained one retains
larger process noise.

The push-dependence reflects a structural asymmetry: only physical pushes
carry kinematic evidence about $(\vartheta_1, \vartheta_2, r)$, while scene
context and directional cues constrain only the goal. We set
$\varsigma_k = 1$ until the first push, freezing the $\mathbf{A}$ matrices at
their scene-prior initialization; once a push arrives, $\varsigma_k$ drops
and process noise becomes active across the shape components.


\noindent\textbf{Directional hints.}\;
Spoken utterances and pointing gestures refine the posterior without
physical contact. Each such observation produces a directional payload $\mathbf{u}_j$
from~\eqref{eq:dir_payload}, where
$\mathbf{d}^{\text{cmd}}_j \in S^1$ is the commanded direction vector
expressed in the robot's heading frame,
$\mathbf{x}_{\text{robot},j}$ is the robot position, and
$\psi_{\text{robot},j} \in S^1$ is the robot heading at command time.
When $\text{source}_j = \text{voice}$, a keyword-based intent classifier
maps the transcript to a fixed direction vector (e.g.\ ``go left''
$\mapsto (0,-1)$); when $\text{source}_j = \text{gesture}$, the direction
is decoded from the user's pointing ray via a hand-pose estimator.

We map the scalar heading to its unit-vector representation
$\mathbf{d}_j = (\cos \psi_{\text{robot},j},\, \sin \psi_{\text{robot},j})$
and write the components of the particle goal $\mathbf{x}^{*(n)}$ as
$(g_x^{(n)},\, g_y^{(n)})$ and the robot position as
$(p_{x,j},\, p_{y,j})$. The cross and dot products of the heading
with the robot-to-goal vector are
\begin{equation}
  \chi_j^{(n)} = d_{x,j}\bigl(g_y^{(n)} - p_{y,j}\bigr)
            - d_{y,j}\bigl(g_x^{(n)} - p_{x,j}\bigr),
  \label{eq:cross}
\end{equation}
\begin{equation}
  \delta_j^{(n)} = d_{x,j}\bigl(g_x^{(n)} - p_{x,j}\bigr)
              + d_{y,j}\bigl(g_y^{(n)} - p_{y,j}\bigr),
  \label{eq:dot}
\end{equation}
from which we obtain the signed bearing angle from the heading to
particle~$n$'s goal:
\begin{equation}
  \alpha_j^{(n)} = \operatorname{atan2}\!\bigl(\chi_j^{(n)},\;
                   \delta_j^{(n)}\bigr).
  \label{eq:bearing}
\end{equation}
The ideal bearing angle for the command is simply the angle of the
commanded direction vector in the heading frame:
\begin{equation}
  \alpha_j^{\mathrm{cmd}} = \operatorname{atan2}\!\bigl(d^{\text{cmd}}_{y,j},\;
                 d^{\text{cmd}}_{x,j}\bigr).
  \label{eq:ideal_angle}
\end{equation}
Because $\mathbf{d}^{\text{cmd}}_j$ is a free unit vector, the
formulation is not restricted to a fixed vocabulary of named commands.
A keyword classifier that maps ``go left'' to $(0,-1)$ or a pointing
gesture decoded as a ray direction both
produce the same directional payload and enter the filter through the same
likelihood. In the experiments reported here we use four voice
commands - straight, right, left, and back, mapped to
$(1,0)$, $(0,1)$, $(0,-1)$, and $(-1,0)$ respectively.

The directional likelihood is
\begin{equation}
  p\!\left(\mathbf{z}^{\text{dir}}_j \mid \boldsymbol{\theta}^{(n)}\right)
  \propto
  \kappa_d^{\,\gamma_j^{(n)}},
  \qquad \kappa_d > 1,
  \label{eq:voice_lik}
\end{equation}
where
\begin{equation}
  \gamma_j^{(n)} = \max\!\left(
    \exp\!\left(-\frac{(\Delta\alpha_j^{(n)})^{2}}
                      {2\,\sigma_d^{2}}\right),\;
    \epsilon_d
  \right),
  \label{eq:gamma}
\end{equation}
and $\Delta\alpha_j^{(n)}$ is the angular difference
$\alpha_j^{(n)} - \alpha_j^{\mathrm{cmd}}$ wrapped to $[-\pi, \pi]$.
When the candidate-goal bearing aligns with the commanded bearing,
$\Delta\alpha_j^{(n)} = 0$ and $\gamma_j^{(n)} = 1$, giving the
particle the full boost $\kappa_d$. The score decreases as a Gaussian
with the angular difference from the commanded bearing, with width
$\sigma_d$, and is bounded below by $\epsilon_d > 0$. Thus, particles
whose candidate goals are poorly aligned with the commanded direction
receive a smaller relative weight rather than being hard-excluded.

Unlike a push, a directional hint has no temporal extent: its payload
$\mathbf{u}_j$ is complete at the instant the command is decoded. It
therefore enters the filter at a single step, and is not reapplied thereafter.

\smallskip
\subsubsection{Resampling (Line~\ref{line:resample})}
\label{sec:resample}

After the weights are normalized, systematic resampling is performed at
every update that carries evidence. A new particle set is drawn from the
current weighted distribution, so particles that better explain the
observations are more likely to be replicated, while low-weight particles
are discarded. Resampling is also what accumulates evidence in this filter. The particle
set entering an update carries uniform weights, so the posterior
$p(\boldsymbol{\theta}_{k-1} \mid \mathbf{z}_{1:k-1})$ is encoded by the
density of particles in parameter space rather than by their weights. To prevent depletion over the discrete goal set, we enforce a floor of m particles per candidate goal after each resampling step, with m small enough that the induced uniform component Km/N is negligible relative to the evidence. Because each particle retains its initial goal assignment, an unfloored filter could extinguish the true goal's cluster after a single misleading cue, with no mechanism for recovery.

\smallskip
\subsubsection{Commitment and Execution (Line~\ref{line:map})}
The filter commits once the current observation is complete: for a push,
after the final increment~\eqref{eq:increment} of the window $\tau$
($\tau = 5\,\mathrm{s}$); for a directional hint, at the step in which it is
decoded.

Because goals are static (Section~\ref{process_model}), the particles
partition into one cluster $\mathcal{N}_i$ per candidate goal. We commit to
the goal whose cluster carries the largest total weight,
$i_k^* = \arg\max_i \sum_{n \in \mathcal{N}_i} w_k^{(n)}$, which is the
posterior probability of that goal: the cluster population records past
evidence, the weights record the current cue. The shape is taken from the
highest-weight particle in that cluster,
$n_k^* = \arg\max_{n \in \mathcal{N}_{i_k^*}} w_k^{(n)}$, giving the LDS policy
\begin{equation}
\label{eq:ds_policy}
\dot{\mathbf{x}} = \mathbf{A}^*(\mathbf{x} - \mathbf{x}^*),
\end{equation}
with $\mathbf{A}^*$ reconstructed via~\eqref{eq:reconstruct}. The robot
follows this policy until a new interaction arrives.

\section{Experiments and results}
In our experiments, we use an SBlimp ~\cite{xu2023sblimp},
a Crazyflie~2.1 quadrotor with tilted rotors rigidly attached to a Mylar
helium balloon. The blimp's pendulum-like stability lets us command
translational velocity directly without an attitude loop, which is precisely
the interface the inferred LDS produces. The system
$\dot{\mathbf{x}}^{d} = \mathbf{A}^{*}(\mathbf{x} - \mathbf{x}^{*})$ is
evaluated at the current position and sent as a desired linear velocity to
the onboard controller~\cite{xu2023sblimp}. The blimp is localized with
an OptiTrack motion-capture system, and
velocity commands are issued through the crazyswarm
framework~\cite{preiss2017crazyswarm} over the crazyradio link.

\noindent\textbf{Inference and interaction pipeline:}\;The proposed particle filter is implemented in NumPy on CPU. Physical
pushes are detected from the motion-capture velocity stream. When the speed
$\|\dot{\mathbf{x}}\|$ exceeds a threshold inconsistent with the current
commanded velocity, the subsequent $\tau$-second window of
$\{\mathbf{x}(t),\, \dot{\mathbf{x}}(t)\}$ is logged as a push observation
$\mathcal{D}_\tau^{(j)}$ in ~\eqref{eq:push_obs}. Voice is captured
through the laptop microphone and decoded with Vosk~\cite{vosk}; gestures
are captured through the laptop webcam and decoded with MediaPipe
Hands~\cite{MediaPipe}. Both run in background threads on the same
workstation, post decoded directional payloads $\mathbf{u}_j$ from \eqref{eq:dir_payload} to a thread-safe queue, and are polled each
frame by the main loop. No part of the pipeline besides the onboard
velocity controller runs on the blimp itself.

\noindent\textbf{Assumptions:}\;Across all scenarios, the robot has no prior
knowledge of obstacles in the environment, while the human knows the obstacle
layout and interacts in a way consistent with avoiding them. For Experiments 1 and 2, the scene
context prior is set to be uniform across the $K$ candidate goals
($\pi_i = 1/K$), so that no goal is favored a priori and the filter's
belief is shaped entirely by the sequence of human interactions. A
trial is counted as a \emph{success} if the robot subsequently reaches
the predesignated goal in less than 6 interactions; otherwise it is a failure. All reported trials use pushes and voice commands only. The gesture channel
is implemented but not evaluated separately, since gestures and voice reduce
to the same directional payload~\eqref{eq:dir_payload} and the same
likelihood~\eqref{eq:voice_lik}.

\begin{figure}[b]
\centering
    \includegraphics[width=0.9\linewidth]{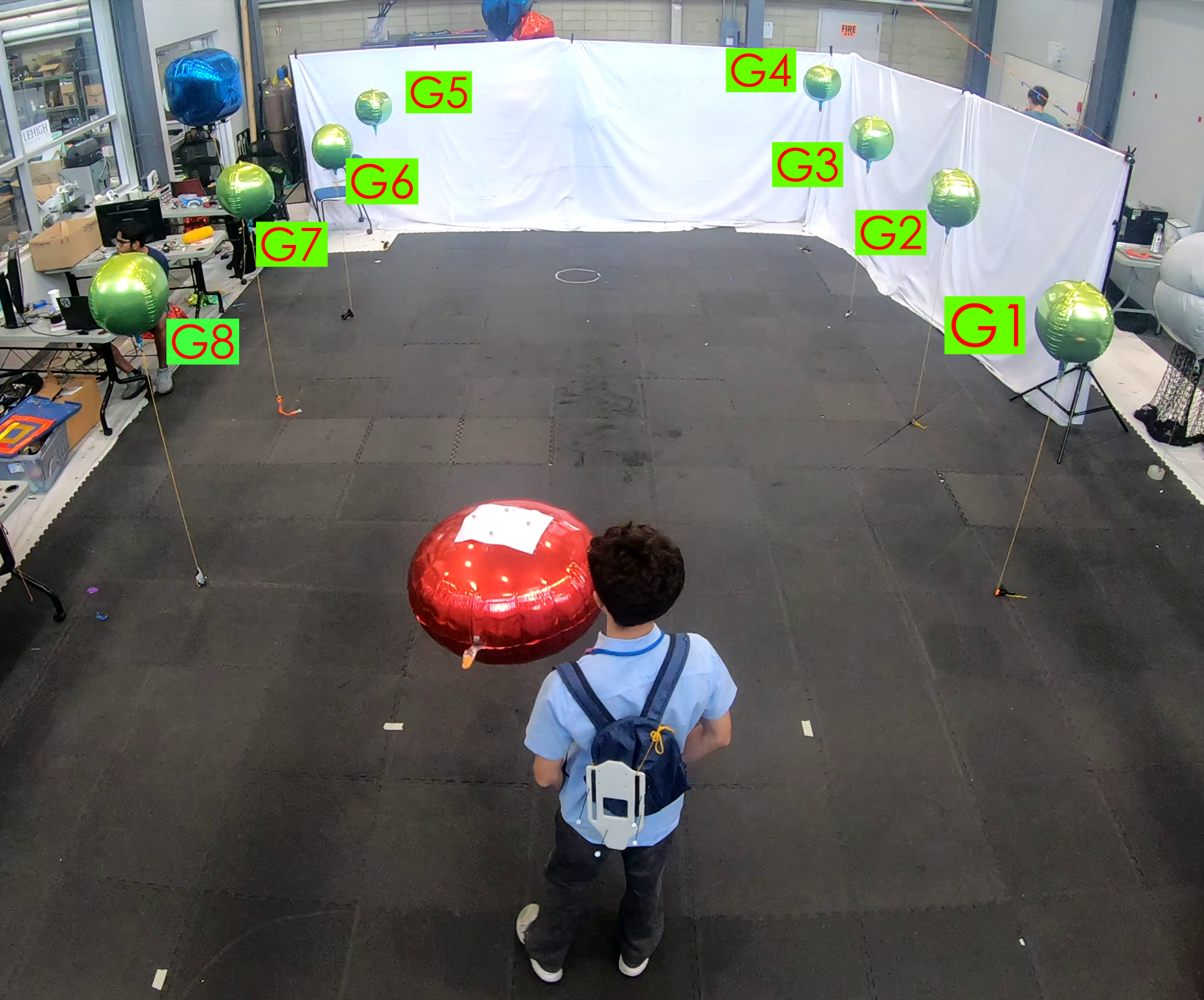}
    \caption{Setup for Experiment 1 where the user has to send the robot to one of the 8 candidate goals selected randomly at the beginning of the run.}
    \label{fig:exp1}
\end{figure}

\subsection{Experiment 1: Goal Identification and Modality}
\label{sec:exp1}

This experiment evaluates how reliably the framework identifies the human's intended goal and quantifies the contribution of each interaction modality. Five participants each completed 60 trials yielding 300 trials in total.

The scene contains $K = 8$ candidate goals
arranged around the workspace (see Fig.~\ref{fig:exp1}). At the
start of each trial, one of the eight goals is selected uniformly at
random as the target, unknown to the robot. The human
then conveys this target through three main ways:
\begin{enumerate}
    \item \textbf{Push only:} the participant may only physically push
          the robot.
    \item \textbf{Voice only:} the participant may only issue
          directional voice commands (e.g.\ ``go left'').
    \item \textbf{Combined:} the participant may freely mix pushes and
          voice commands.
\end{enumerate}

\begin{figure}[ht]
    \centering
    \includegraphics[width=1.0\linewidth]{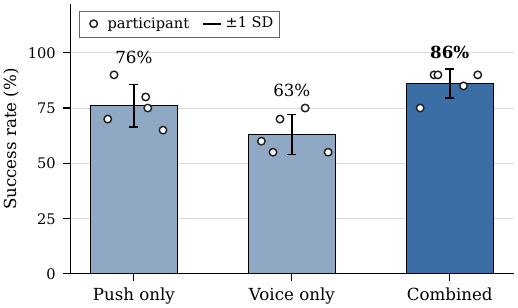}
    \caption{Success rate by interaction modality, pooled across 100 trials per condition (five participants, 20 trials each). Open circles show individual participant rates.}
    \label{fig:modality_bar2}
\end{figure}

\begin{figure}[ht]
    \centering
    \includegraphics[width=1.0\linewidth]{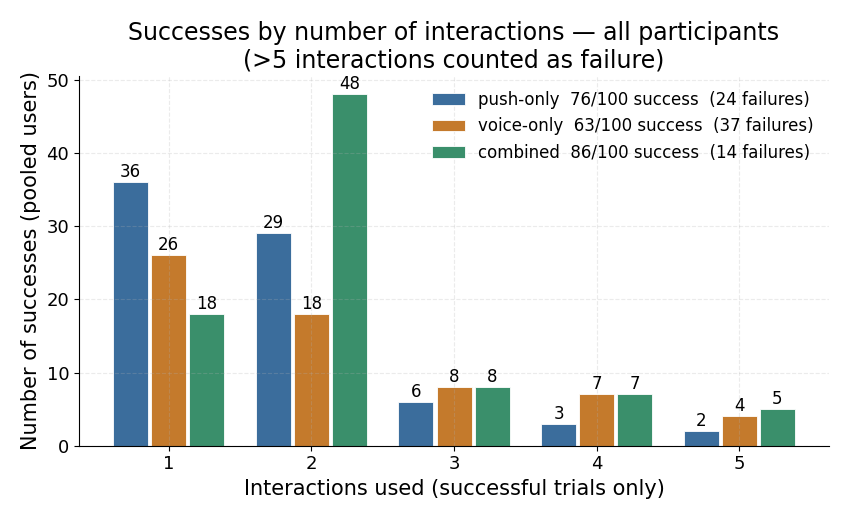}
    \caption{Distribution of the number of interactions used in successful trials, pooled across all five participants. Each participant completed 20 trials per modality, giving 100 trials per modality across the five participants}
    \label{fig:num_interactions}
\end{figure}

Figures \ref{fig:modality_bar2} and \ref{fig:num_interactions} show that the three modalities succeed through different interaction patterns. Push-only is a high-information channel, nearly half of its successes are resolved in a single interaction, but when that first push is ambiguous the participant has to physically go to the robot to correct its trajectory, which increases human effort. Voice-only provides weaker directional evidence per interaction, spreading its successes across one to five commands and producing the highest failure rate (37/100). The combined condition resolves this issue: successes concentrate sharply at two interactions, consistent with a two-stage strategy in which a push establishes coarse direction and a voice command disambiguates among the remaining candidates. 

After completing all three conditions, participants rated the ease of conveying their goal and their confidence in the robot's understanding on a 5-point Likert scale, and answered free-response questions about their strategies and difficult goal locations. All five participants rated both ease and confidence at 4 or above, and four preferred the combined modality. All participants independently reported using a two-stage ``push-then-voice'' strategy, with an initial push providing coarse directional information and a subsequent voice command disambiguating among the remaining goals. This is consistent with the two-interaction peak observed in Figure \ref{fig:num_interactions}. Participants also identified the central targets, particularly Goals~3 and~6, as the most difficult to communicate because of their proximity to neighboring goals.

\subsection{Experiment 2: Curved Trajectories Around Obstacles}
We tested whether the framework can infer curved trajectories that avoid obstacles, using a scene with 3 goals and a single obstacle. The participant's task was to guide the robot to a goal whose straight-line path was blocked, using any combination of pushes and voice commands. Because the eigenvalue parameterization in Section~\ref{methodology}
admits admits a broad class of stable-node dynamics, the inferred dynamical system is not
restricted to straight-line convergence.
This lets the human steer the robot around obstacles known only to
them, without ever communicating the obstacle location explicitly.

Fig.~\ref{fig:comp_tile.png} illustrates this on a scene in which the human
wants the robot to reach Goal~2, but a straight-line trajectory would
collide with the obstacle. Rather than pushing toward Goal~2 directly,
the human pushes the robot toward Goal~1. After this first push, the filter retains hypotheses consistent with
all the goals, with Goal~1 most heavily weighted because the push
aligns most closely with the straight-line direction to it. The human
then issues the voice command ``go right,'' which reweights particles
whose goals lie to the right of the robot's heading. The filter commits to Goal~2, and the executed
trajectory curves around the obstacle.

\subsection{Experiment 3: Importance of priors}

\begin{figure}
    \centering
    \includegraphics[width=1.0\linewidth]{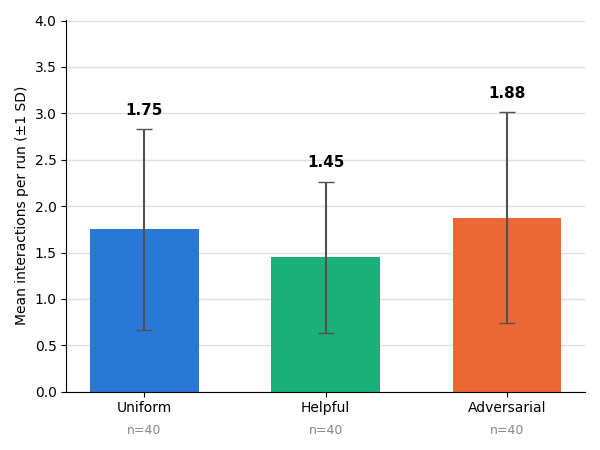}
    \caption{Interactions to successful commitment under three prior
    conditions ($n=40$ each). The adversarial prior, despite concentrating
    40\% of particles on the farthest goal, matches the uniform baseline,
    while the helpful prior shows a modest reduction. 
    }
    \label{fig:prior}
\end{figure}

To evaluate sensitivity to the prior distribution, we repeated the combined condition of Experiment~1 under three initializations: uniform (particles distributed equally across all goals), helpful (40\% of particles concentrated on the true target and the remainder distributed uniformly across the other goals), and adversarial (40\% of particles concentrated on the goal farthest from the true target and the remainder distributed uniformly across the other goals). To isolate the prior from sensor and actuation noise, these trials were run in simulation using similar filter configuration as Experiments~1 and~2; absolute success rates and convergence speeds may therefore differ in the real world. Each
condition comprised $n=40$ trials, and we report the number of
interactions required to commit to the correct goal (Fig.~\ref{fig:prior}).

The results show that our framework is robust to priors. Even with 40\% of its mass on the farthest goal, the adversarial prior needed about as many interactions
as the uniform baseline ($1.88$ vs.\ $1.75$), showing that one informative
cue is enough to overcome a bad initialization. This is because our filter
is evidence-dominated, and since we resample at every
step, the first cue reshapes the particle cloud very fast and largely overwrites the prior. The helpful prior did somewhat better ($1.45$), but the effect is
small and hence we do not claim it as a strong result. 


\section{Conclusion}

We presented a novel multimodal intent-inference framework that represents human intent as a linear dynamical system and estimates its parameters online using a particle filter. Physical pushes and directional hints enter the filter as complementary sources of evidence through modality-specific likelihoods, enabling the robot to accumulate partial cues over time rather than requiring any single interaction to fully specify the goal or behavior. Experiments on a physical blimp showed that combining pushes and voice commands improves on either alone, and that the eigenvalue parameterization lets the filter infer curved trajectories around obstacles known only to the human.


For future work, we will scale the study to more participants in complex environments. We also want to extend the modalities to natural language, body language and facial expressions for different applications.


\bibliographystyle{IEEEtran}
\bibliography{ref.bib}
\end{document}